# Explainable Modeling for Wind Power Forecasting: A Glass-Box Approach with High Accuracy

Wenlong Liao, Fernando Porté-Agel, Jiannong Fang, Birgitte Bak-Jensen, Guangchun Ruan, Zhe Yang

*Abstract*—Machine learning models (e.g., neural networks) achieve high accuracy in wind power forecasting, but they are usually regarded as black boxes that lack interpretability. To address this issue, the paper proposes a glass-box approach that combines high accuracy with transparency for wind power forecasting. Specifically, the core is to sum up the feature effects by constructing shape functions, which effectively map the intricate non-linear relationships between wind power output and input features. Furthermore, the forecasting model is enriched by incorporating interaction terms that adeptly capture interdependencies and synergies among the input features. The additive nature of the proposed glass-box approach ensures its interpretability. Simulation results show that the proposed glass-box approach effectively interprets the results of wind power forecasting from both global and instance perspectives. Besides, it outperforms most benchmark models and exhibits comparable performance to the best-performing neural networks. This dual strength of transparency and high accuracy positions the proposed glass-box approach as a compelling choice for reliable wind power forecasting.

## I. Introduction

IN light of the energy crisis and the urgency to accelerate carbon neutrality, the effective utilization of wind power generation has emerged as a prominent topic of interest. Wind power stands out as a promising renewable energy source, and its integration into smart grids is steadily on the rise [1]. Despite the favorable environmental and economic impacts attributed to wind power generation, the intermittent and fluctuating nature of wind speed (WS) presents a significant challenge in achieving highly accurate wind power forecasting [2]. This challenge, in turn, poses risks to the stable operation of smart grids [3]. As a result, the development of accurate and transparent wind power forecasting becomes imperative to facilitate the safe operation and planning of smart grids.

Definitions of time horizons for wind power forecasting vary slightly from publication to publication [4], [5]. For example, [6] categorizes wind power forecasting into long-term (from weeks to months), medium-term (from days to weeks), short-term (from hours to days), and very short-term (from seconds to hours). Within this paper, the focus is primarily on short-term wind power forecasting, according to the above definition.

Most previous publications fall into three main categories [6]: physical methods, statistical methods, and artificial intelligence (AI) methods. Characterized by detailed physical modeling of wind turbines, physical methods often involve extensive physical parameters [7], such as numerical weather prediction (NWP) data, obstacles, surface roughness, and terrain. Complex mathematical models integrate these inputs to estimate wind speeds that are matched to wind turbine power curves for power forecasting. While this approach requires no historical data, it relies on a large amount of physical information and is computationally intensive.

Statistical methods typically utilize historical wind power data to forecast future values. Classical examples include the persistence model (PM), gray forecasting model, autoregressive model, and autoregressive moving average model [8], [9]. While statistical methods are cost-efficient, their forecasting accuracy tends to decrease as the forecasting time horizon extends.

Normally, AI methods map input variables (e.g., historical wind power data or NWP data) to wind power output through supervised learning. Traditional AI methods, such as regression tree (RT) [10], linear regression (LR) [11], generalized additive model (GAM) [12], and ridge regression, manifest as interpretable glass-box models. For instance, LR establishes a linear relationship between input features and wind power output in [13], making it easy to interpret and analyze. Typically, these traditional AI methods are amenable to understanding, facilitating insight into the impact of features, and have modest data requirements. However, their accuracy diminishes when faced with complex relationships, as they struggle to encapsulate non-linear patterns, resulting in limited forecasting accuracy. To improve the forecasting accuracy, more recent AI methods, such as multi-layer perception (MLP) [14], support vector regression (SVR) [15], extreme gradient

This work is funded by the Swiss Federal Office of Energy (Grant No. SI/502135–01). Also, this work is carried out in the frame of the "UrbanTwin: An urban digital twin for climate action: Assessing policies and solutions for energy, water and infrastructure" project with the financial support of the ETH-Domain Joint Initiative program in the Strategic Area Energy, Climate and Sustainable Environment.

W. Liao F. Porté-Agel, and J. Fang are with Wind Engineering and Renewable Energy Laboratory, Ecole Polytechnique Federale de Lausanne (EPFL), Lausanne 1015, Switzerland.
B. Bak-Jensen is with the AAU Energy, Aalborg University, Aalborg 9220, Denmark.
G. Ruan is with the Laboratory For Information & Decision Systems, Massachusetts Institute of Technology, Cambridge 02139, United States.
Z. Yang (corresponding author) is with the Department of Electrical Engineering, The Hong Kong Polytechnic University, Hong Kong (e-mail: zhe1yang@polyu.edu.hk).

boosting (XGBoost) [16], and long short-term memory (LSTM) [17], have been proposed to capture latent features and non-linear patterns. In contrast to traditional AI methods, these more contemporary ones offer enhanced forecasting capabilities, but they are perceived as black-box models, which hinders the interpretation of the decision-making process.

As an example, Fig.1 provides a visual insight into the accuracy and interpretability of five popular AI methods [18], [19], [20]. Note that this figure is indicative of the majority of datasets, but there may be variations in accuracy rankings for specific datasets. It is clear that most existing AI methods suffer from certain limitations.
- Traditional AI methods (e.g., LR and RT) are amenable to understanding, but their forecasting accuracy is limited.
- Although more recent AI methods (e.g., XGBoost and neural network) perform quite well, they are black-box models (i.e., hard to interpret).
- There is a need to develop a glass-box model with high accuracy.

In light of these limitations, this paper introduces a novel approach by proposing the construction of an innovative glass-box model, aiming to combine accuracy and interpretability for wind power forecasting. The key contributions are as follows:
- By employing advanced AI models (e.g., gradient boosting) to construct shape functions and incorporating interaction terms between input features in the wind power forecasting model, this paper proposes a new approach for wind power forecasting, which holds the dual strength of transparency and high accuracy.
- A unique dimension is presented to unravel the intricate relationship between input features and wind power output from two perspectives. A global perspective reveals the overarching effects of input features and their interaction terms, aiding in the construction of feature engineering. Meanwhile, an instance perspective delves into individual contributions, which helps to identify and correct errors, thereby improving the accuracy and reliability of wind power forecasting.
- A comprehensive case study is conducted on seven wind power datasets to validate and analyze the accuracy, time efficiency, and interpretability of the proposed glass-box approach compared to commonly used black-box models (e.g., XGBoost, SVR, MLP, and LSTM) and glass-box models (e.g., LR, RT, PM, and GAM).

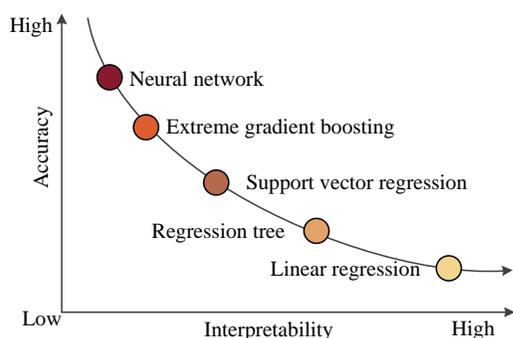
Fig. 1. The accuracy and interpretability of five popular AI methods in wind power forecasting [18].

The rest of the paper is organized as follows. Section II shows the wind power forecasting model and explains why interpretability is needed. Section III formulates the proposed novel glass-box model. A comprehensive case study is conducted in section IV. Section V presents the conclusions.

## II. PROBLEM DESCRIPTION

### A. Wind Power Forecasting Model

According to the findings of related publications, future values of wind power can be estimated by using NWP data (e.g., the forecasted values of wind speeds and wind directions from the weather model) and historical wind power observations as input features. Therefore, the wind power forecasting model can be formulated as:

$$y = f(x_1, x_2, \ldots x_n) \qquad (1)$$

where $f()$ is an AI method; $x_i$ is an input feature; $y$ is the future value of wind power; and $n$ is the number of input features.

Usually, when the forecasting time horizon (i.e., look-ahead time) is within a few hours (typically around 4 hours), the input features generally do not require NWP data [4],[5], as historical wind power observations alone can accurately estimate future values. Conversely, when the forecasting time horizon exceeds a few hours, future values are generally estimated by using NWP data as input features [5].

### B. Why is Interpretability Needed?

Fig.2 summarizes the main reasons why interpretability is needed in wind power forecasting [21].

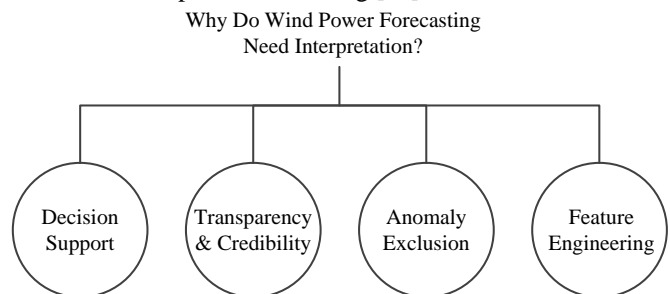
Fig. 2. The reasons why interpretability is needed in wind power forecasting.

**Decision Support**: Interpretability makes wind power forecasting easier to understand, and helps the power system operator (PSO) make more informed operational decisions about wind power integration in smart grids. In addition, the PSO can better cope with the wind power fluctuation by understanding how the glass-box model takes into account different input features and makes forecasts.

**Transparency and Credibility**: Interpretability reveals the functionalities of the glass-box model and enables stakeholders to understand how wind power forecasted values are obtained. This helps build confidence in the forecasting results and reduces misunderstandings.

**Anomaly Exclusion**: If there are outliers in the forecasted values, an interpretable glass-box model can help identify the root cause of the problem. This also helps to identify and correct errors, improving the accuracy and reliability of wind power forecasting.

**Feature Engineering**: The glass-box model can help construct interpretable feature engineering (i.e., understanding the contribution of input features to forecasted values), and transfer the feature engineering to wind power forecasting tasks in other regions.

## III. METHODOLOGY

The objective of this paper is to develop an accurate wind power forecasting model with interpretability, which consists of two main components: the shape function and the interaction term. In particular, shape functions capture the intricate and non-linear relationship between wind power output and input features, while interaction terms represent the interactions between input features.

### A. Shape Functions Capture Non-linear Relationship

Interpretability is the ability to understand the contributions of input features to forecasted values. This takes into account the intricate relationship between individual input features and wind power forecasted values.

Traditionally, wind power can be forecasted by a simple but interpretable linear regression model:

$$y = \alpha_0 + \alpha_1 x_1 + \alpha_2 x_2 + \ldots + \alpha_n x_n \quad (2)$$

where $\alpha_0$ is the intercept; and $\alpha_i$ is the feature weight, representing the contribution of the input feature $x_i$ to the forecasted value.

The linearity of Eq. (2) makes it easy to interpret wind power forecasting, but its forecasting accuracy is limited. The reason behind this is the intricate and non-linear relationship between wind power output and input features.

Therefore, the first contribution of this paper is to map the non-linear relationship by adopting advanced AI methods rather than the linear term $\alpha_i x_i$. The specific model can be formulated as:

$$y = \alpha_0 + f_1(x_1) + f_2(x_2) + \ldots + f_n(x_n) \quad (3)$$

where $f_i()$ is an AI model (also called shape function).

Compared with the linear regression model in Eq.(2), the Eq.(3) replaces the linear term $\alpha_i x_i$ with a more flexible shape function. Obviously, from Eq. (3), it is easy to interpret the wind power forecasting, as the relationship between the input feature $x_i$ and the individual univariate term $f_i(x_i)$ can be visualized by plotting shape functions with $f_i(x_i)$ on the y-axis and $x_i$ on the x-axis.

In Eq.(3), the key point is to construct the shape function to learn the non-linear relationship. In previous publications [19],[20], splines are considered as shape functions, and the linear model is generalized into spline regression model. However, these spline regression models require pre-specifying the shape and number of splines, which can involve subjective choices and may not fully capture nonlinear relationships. In contrast, RT models can more adaptively discover nonlinear relationships without assuming specific spline shapes. Given the flexibility and better adaptability of decision tree models, this paper uses the RT models to construct the shape function.

The RT model groups the data by recursively splitting it based on specific feature thresholds, generating subsets known as terminal or leaf nodes, each containing instances grouped by their features. The non-terminal subsets, referred to as internal or split nodes, serve as intermediate partitions. Predictions are made by averaging the outcomes of the training data within each leaf node. These trees are versatile, applicable to both classification and regression tasks.

There are many tree induction algorithms that differ in factors such as tree structure, splitting criteria, stopping conditions, and leaf node model estimation. Among them, the classification and regression trees (CART) algorithm stands out as one of the most widely used for tree induction. Therefore, CART is employed to construct tree model. Note that while our focus is on CART, the principles of interpretation apply similarly to other tree algorithms. The relationship between the wind power and input features in CART can be formulated as:

$$y = \sum_{m=1}^{M} c_m I\{x \in R_m\} \quad (4)$$

where $c_m$ denotes the predicted value at each leaf node $m$; $I$ is the indicator function, which determines whether the feature $x$ belongs to the region represented by leaf node $m$; and $M$ is the number of the number of leaf nodes.

Eq.(4) essentially describes the process of assigning features to different leaf nodes, calculating the forecasted values at each leaf node, and summing up these forecasted values to obtain the final forecasted wind power.

### B. Interaction Terms Represent Correlations

There is still a significant gap in accuracy between Eq. (3) and complex black-box models (e.g., neural network), since Eq. (3) ignores the interactions between input features.

For instance, Fig. 3 presents the input features used in the wind power forecasting competition (forecasting horizon is 24 hours) from [22]. It is evident that interactions exist among these input features, including U and V at different heights. In particular, the U10 and U100 show the interrelated behavior, while there is also a correlation between the V10 and V100.

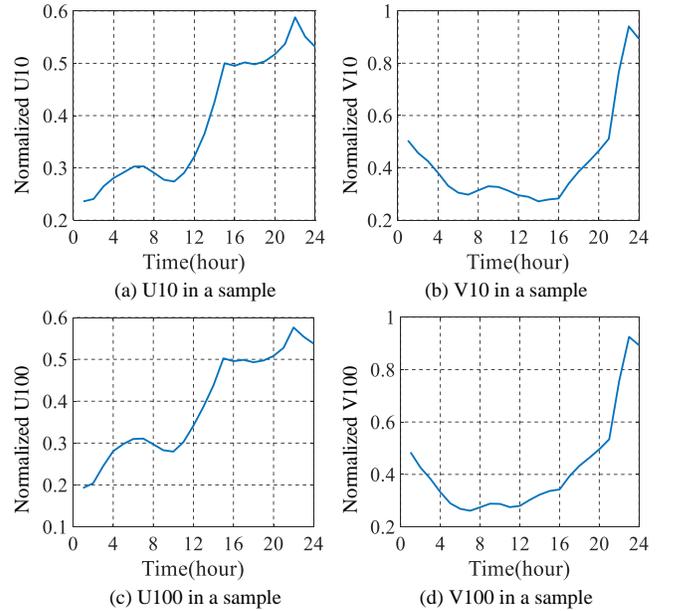

(a) U10 in a sample  (b) V10 in a sample
(c) U100 in a sample  (d) V100 in a sample

Fig. 3. The normalized U and V at different heights in a sample. U10 represents the wind speed at 10 meters in an east-west direction, with positive values indicating wind blowing from west to east. V10 represents the wind speed at 10 meters in a north-south direction, with positive values indicating wind blowing from south to north. U100 and V100 represent the vertical component of wind speeds at a height of 100 meters.

Further, the Pearson correlation coefficients between input features and the wind power are presented in Fig. 4, which quantifies the strength of these interactions. There is a certain degree of correlation between the input features as well as between the input features and the wind power. This underscores the importance of considering interactions between input features when designing wind power forecasting models. Similarly, for the wind power forecasting within a few hours, the input features are typically historical observations, which are also highly correlated [1], [5].

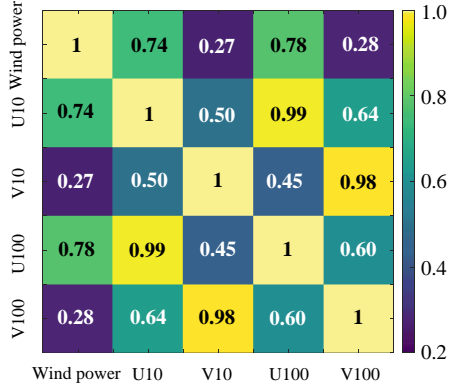

Fig. 4. The Pearson correlation coefficients between input features and the wind power.

Therefore, the second contribution of this paper is to add the interaction terms between input features into the wind power forecasting model:

$$y = \alpha_0 + \sum_{i=1}^{n} f_i(x_i) + \sum_{i=1}^{n-1}\sum_{j=i+1}^{n} f_{i,j}(x_i, x_j) \quad (5)$$

where $f_{i,j}(x_i,x_j)$ is the interaction term between the input feature $x_i$ and input feature $x_j$. Also, the shape function $f_{i,j}()$ and $f_i()$ are also mapped by advanced AI methods (i.e., RT model).

Note that the representation of 2-D interactions can take the form of heat maps that plot $f_{i,j}(x_i,x_j)$ over the 2-D plane. Consequently, the wind power forecasting model containing only 1-D terms and 2-D terms in Eq. (5) remains interpretable.

The proposed novel glass-box approach possesses a remarkable level of interpretability, as it enables the visualization and comprehension of the impact of input features and interaction terms on the forecasted value through graphical representations. Given the additive nature of the proposed glass-box approach, the roles of input features and their interaction terms in wind power forecasting model follow a modular pattern, making it easy to measure their contributions to the forecasted values [20].

For a given sample of wind power forecasting, each shape function serves as a feature-specific lookup table, providing a term contribution. The summation of these term contributions yields the ultimate forecasted values. The additive nature allows the term contributions to be sorted and visualized, revealing the prominent influence of specific input features on each individual wind power forecasting. This visual representation helps to identify the relative impact of various input features on the forecasting process.

## IV. CASE STUDY

### A. Wind Power Dataset Description

To test the proposed glass-box model, simulations are conducted on seven real wind power datasets derived from GEFCom 2014 [22], the national renewable energy laboratory (NREL) [23], and the JUVENT wind farm. Table I briefly summarizes the parameters of these datasets.

TABLE I
WIND POWER DATASET DESCRIPTION

| Datasets | Time span | Forecasting horizon | Input features | Dataset Type |
|---|---|---|---|---|
| GEFCom 2014 | Jan. 2012 to Dec. 2013 | 24 hours | NWP data | Public |
| NREL | Jan. 2012 to Dec. 2012 | 30 minutes to 4 hours | Historical wind powers | Public |
| JUVENT | Jan. 2014 to Apr. 2019 | 30 minutes to 4 hours | Historical wind powers | Private |

GEFCom 2014 [22]: GEFCom 2014 consists of 10 public wind power datasets sourced from 10 zones. These datasets range from January 1, 2012 to December 1, 2013. The time resolution is 1 hour. The input features are NWP data, including U10, V10, U100, and V100. The objective is to forecast wind power generation 24 hours ahead. Here, 5 datasets are randomly selected for simulations.

NREL [23]: The time span of the NREL-derived wind power dataset is from January 1, 2012 to December 31, 2012. This is also a public dataset. The time resolution is 15 minutes after data cleaning. The input features are historical wind powers without NWP data. The objective is to forecast wind power generation 30 minutes to 4 hours ahead.

JUVENT: The JUVENT wind farm is located in the Jura Mountains of Switzerland. The time span of the JUVENT-derived wind power dataset is from January 1, 2014 to April 30, 2019. This is a private wind power dataset. The time resolution is 30 minutes after data cleaning. The input features are also historical wind power without NWP data. The objective is to forecast wind power generation 30 minutes to 4 hours ahead.

The focus of this paper is not on feature selection. Therefore, for the NREL and JUVENT datasets, the lengths of the input features are empirically set to 48 and 24, respectively. In other words, the historical data from the previous 12 hours is used to forecast future values. Of course, hyper-parameter optimization can be used in the future to obtain the optimal length of input features for each case [24], but fairness is maintained across all models due to the uniform use of the same input features.

### B. Evaluation Metrics

For each dataset, the first 80% is used as a training set, and the next 10% as a validation set. The last 10% is used as a test set to evaluate the model performance by widely used metrics, including normalized root mean square error (NRMSE), normalized mean absolute error (NMAE), and the coefficient of determination ($R^2$):

$$NRMSE = \sqrt{\frac{1}{m}\sum_{i=1}^{m}(\hat{y}_i - y_i)^2} \quad (6)$$

$$NMAE = \frac{1}{m}\sum_{i=1}^{m}|\hat{y}_i - y_i| \quad (7)$$

$$R^2 = 1 - \frac{\sum_{i=1}^{m}(\hat{y}_i - y_i)^2}{\sum_{i=1}^{m}(y_i - \bar{y})^2} \tag{8}$$

where $m$ is the number of forecasting points; $y$ is the normalized true value; $\hat{y}_i$ is the normalized forecasted value; and $\bar{y}$ is the mean value of all observations. The closer NRMSE and NMAE are to 0, the better the model performance. Conversely, the closer $R^2$ is to 1, the better the model performance.

*C. Benchmark Models*

To test the model performance, the proposed glass-box approach, i.e., explainable boosting machine for wind power forecasting (denoted as WindEBM), will be compared with commonly used methods, including black-box models (e.g., XGBoost, SVR, MLP, and LSTM) and glass-box models (e.g., LR, RT, PM, and GAM).

All algorithms are executed on a Spyder with AI libraries, including Tensorflow 2.0 and InterpretML 0.4. The parameters of the computer are as follows: Intel(R) Core(TM) i5-10210U CPU@1.60GHz,2.11 GHz; RAM 8GB.

Hyper-parameter optimization and cross-validation [24] are employed to determine the optimal parameters of each method. For instance, the hyper-parameters of each method used to forecast wind power 30 minutes ahead on the JUVENT dataset are presented below.

**XGBoost** [16]: The number of estimators is 500. The min sample split is 5. The maximum depth of the tree is 4. The learning rate is 0.001. The loss function is the squared error.

**SVR** [15]: The kernel is linear. The early stopping tolerance is 0.0001.

**MLP** [14]: It consists of four dense layers with neuron counts of 64, 32, 16, and 1, respectively. The activation function of the last layer is sigmoid, while the previous layers use the rectified linear unit (ReLU) activation. The learning rate is set to 0.01, and the optimizer is the Adam algorithm.

**LSTM** [17]: It consists of two LSTM layers and two dense layers. They have 64, 32, 16, and 1 neurons, respectively. The other parameters of the LSTM are the same as those of the MLP.

**LR** [11]: There are no hyper-parameters.

**RT** [10]: The max depth of the tree is 4. The min sample split is 4, and the min sample leaf is 1. The criterion for splitting is NMAE.

**PM**: There are no hyper-parameters.

**GAM** [12]: The shape function is the spline. Each feature is set to a linear smoothing term.

**WindEBM**: learning rate is 0.001, and the early stopping tolerance is 0.0001. The min sample split is 5. The max round is 5000. The shape functions are constructed by using the RT model. The parameters are the same as described earlier.

*D. Performance Comparison with Benchmark Models*

To evaluate the performance of the proposed WindEBM and benchmark models, experiments are conducted on seven real datasets. Each method is independently run 50 times, and the average evaluation metrics are presented in Tables II-IV. Note that PM is suitable for wind power forecasting within a few hours, but not for forecasting 24 hours ahead. Thus, PM is excluded from Table II.

From Table II, the proposed WindEBM with interpretability emerges as a standout performer, outperforming other glass-box approaches (e.g., LR, RT, PM, and GAM). For example, the proposed WindEBM achieves remarkable forecasting accuracy across all evaluation metrics, with an average $R^2$ of 0.713 on dataset 1 from the GEFCOM2014. Impressively, the proposed WindEBM also outperforms most black-box models (e.g., XGBoost, SVR, and MLP), and it is comparable to the top-performing neural network (i.e., LSTM).

Note that GAM is a simplified version of WindEBM (i.e., Eq. (3) with splines used as shape functions), and the comparison between GAM and WindEBM belongs to the ablation experiment, demonstrating that the proposed two contributions of this paper (i.e., the use of gradient boosting to construct shape functions and the introduction of interaction terms) significantly improve the model performance.

To visualize the forecasted results, we randomly selected a subset of forecasts from dataset 1 in Table II, as shown in Fig. 5. The dashed lines represent the forecasted values from traditional glass-box models, while the solid lines represent those from black-box models.

From Fig. 5, it can be observed that the forecasted accuracy of these traditional glass-box model (e.g., LR, RT, PM, and GAM) is limited. However, the proposed WindEBM demonstrates forecasted accuracy comparable to advanced black-box models. This further validates the findings presented in Table II.

From Table III and Table IV, the proposed WindEBM presents an excellent performance for wind power forecasting within a few hours, which holds true across datasets from diverse geographical regions (e.g., NREL and JUVENT). For example, when forecasting wind power 4 hours ahead using the JUVENT dataset, the competitive advantage of WindEBM is evident when compared to benchmark models. WindEBM consistently delivers robust results that rival the best performers, as evidenced by its average NRMSE, NMAE, and $R^2$ values of 0.186, 0.133, and 0.593, respectively.

TABLE II
RESULTS OF WIND POWER FORECASTING 24 HOURS AHEAD ON DATASETS FROM THE GEFCOM 2014

| Method | Dataset 1, GEFCom 2014 | | | Dataset 2, GEFCom 2014 | | | Dataset 3, GEFCom 2014 | | | Dataset 4, GEFCom 2014 | | | Dataset 5, GEFCom 2014 | | |
|---|---|---|---|---|---|---|---|---|---|---|---|---|---|---|---|
| | NRMSE | NMAE | $R^2$ | NRMSE | NMAE | $R^2$ | NRMSE | NMAE | $R^2$ | NRMSE | NMAE | $R^2$ | NRMSE | NMAE | $R^2$ |
| WindEBM | 0.182 | 0.135 | 0.713 | 0.156 | 0.107 | 0.706 | 0.169 | 0.114 | 0.628 | 0.130 | 0.093 | 0.765 | 0.174 | 0.126 | 0.737 |
| LR | 0.335 | 0.297 | 0.028 | 0.262 | 0.220 | 0.166 | 0.260 | 0.218 | 0.123 | 0.247 | 0.206 | 0.150 | 0.313 | 0.273 | 0.147 |
| RT | 0.259 | 0.213 | 0.420 | 0.201 | 0.159 | 0.510 | 0.211 | 0.165 | 0.424 | 0.186 | 0.147 | 0.518 | 0.241 | 0.194 | 0.493 |
| GAM | 0.194 | 0.150 | 0.674 | 0.163 | 0.114 | 0.679 | 0.175 | 0.121 | 0.603 | 0.140 | 0.102 | 0.728 | 0.195 | 0.147 | 0.670 |
| XGBoost | 0.184 | 0.139 | 0.708 | 0.158 | 0.111 | 0.702 | 0.171 | 0.119 | 0.624 | 0.133 | 0.097 | 0.753 | 0.178 | 0.132 | 0.724 |
| SVR | 0.215 | 0.159 | 0.599 | 0.169 | 0.120 | 0.653 | 0.179 | 0.122 | 0.583 | 0.140 | 0.103 | 0.728 | 0.193 | 0.146 | 0.676 |
| LSTM | 0.179 | 0.131 | 0.724 | 0.155 | 0.108 | 0.710 | 0.166 | 0.114 | 0.631 | 0.130 | 0.093 | 0.765 | 0.173 | 0.124 | 0.740 |
| MLP | 0.183 | 0.137 | 0.711 | 0.157 | 0.110 | 0.699 | 0.170 | 0.116 | 0.625 | 0.132 | 0.094 | 0.761 | 0.177 | 0.130 | 0.728 |

TABLE III
RESULTS OF WIND POWER FORECASTING A FEW HOURS AHEAD ON DATASETS FROM THE NREL

| Method | Forecasting horizon is 0.5 h | | | Forecasting horizon is 1 h | | | Forecasting horizon is 2 h | | | Forecasting horizon is 3 h | | | Forecasting horizon is 4 h | | |
|---|---|---|---|---|---|---|---|---|---|---|---|---|---|---|---|
| | NRMSE | NMAE | $R^2$ | NRMSE | NMAE | $R^2$ | NRMSE | NMAE | $R^2$ | NRMSE | NMAE | $R^2$ | NRMSE | NMAE | $R^2$ |
| WindEBM | 0.072 | 0.034 | 0.969 | 0.088 | 0.050 | 0.954 | 0.137 | 0.083 | 0.889 | 0.171 | 0.112 | 0.824 | 0.186 | 0.121 | 0.793 |
| LR | 0.076 | 0.035 | 0.965 | 0.092 | 0.052 | 0.950 | 0.147 | 0.089 | 0.871 | 0.178 | 0.118 | 0.810 | 0.207 | 0.145 | 0.745 |
| RT | 0.082 | 0.044 | 0.959 | 0.097 | 0.057 | 0.944 | 0.151 | 0.092 | 0.865 | 0.184 | 0.121 | 0.798 | 0.211 | 0.147 | 0.738 |
| GAM | 0.077 | 0.036 | 0.964 | 0.093 | 0.051 | 0.948 | 0.148 | 0.088 | 0.869 | 0.179 | 0.118 | 0.807 | 0.209 | 0.142 | 0.742 |
| PM | 0.077 | 0.034 | 0.965 | 0.093 | 0.051 | 0.949 | 0.150 | 0.085 | 0.867 | 0.186 | 0.115 | 0.792 | 0.217 | 0.124 | 0.721 |
| XGBoost | 0.078 | 0.040 | 0.964 | 0.092 | 0.053 | 0.949 | 0.141 | 0.086 | 0.882 | 0.176 | 0.115 | 0.813 | 0.202 | 0.138 | 0.758 |
| SVR | 0.097 | 0.069 | 0.943 | 0.112 | 0.080 | 0.925 | 0.154 | 0.108 | 0.860 | 0.199 | 0.137 | 0.762 | 0.221 | 0.153 | 0.710 |
| LSTM | 0.072 | 0.035 | 0.968 | 0.086 | 0.050 | 0.956 | 0.134 | 0.081 | 0.894 | 0.168 | 0.110 | 0.834 | 0.181 | 0.119 | 0.806 |
| MLP | 0.075 | 0.038 | 0.966 | 0.091 | 0.054 | 0.951 | 0.138 | 0.086 | 0.887 | 0.175 | 0.114 | 0.816 | 0.197 | 0.136 | 0.771 |

TABLE IV
RESULTS OF WIND POWER FORECASTING A FEW HOURS AHEAD ON DATASETS FROM THE JUVENT

| Method | Forecasting horizon is 0.5 h | | | Forecasting horizon is 1 h | | | Forecasting horizon is 2 h | | | Forecasting horizon is 3 h | | | Forecasting horizon is 4 h | | |
|---|---|---|---|---|---|---|---|---|---|---|---|---|---|---|---|
| | NRMSE | NMAE | $R^2$ | NRMSE | NMAE | $R^2$ | NRMSE | NMAE | $R^2$ | NRMSE | NMAE | $R^2$ | NRMSE | NMAE | $R^2$ |
| WindEBM | 0.111 | 0.069 | 0.856 | 0.127 | 0.083 | 0.806 | 0.160 | 0.107 | 0.709 | 0.177 | 0.124 | 0.651 | 0.186 | 0.133 | 0.593 |
| LR | 0.111 | 0.071 | 0.855 | 0.130 | 0.083 | 0.798 | 0.162 | 0.110 | 0.694 | 0.180 | 0.130 | 0.630 | 0.201 | 0.144 | 0.513 |
| RT | 0.118 | 0.076 | 0.837 | 0.133 | 0.088 | 0.788 | 0.165 | 0.110 | 0.686 | 0.183 | 0.130 | 0.619 | 0.209 | 0.150 | 0.476 |
| GAM | 0.113 | 0.075 | 0.849 | 0.139 | 0.091 | 0.767 | 0.171 | 0.115 | 0.660 | 0.190 | 0.136 | 0.589 | 0.213 | 0.154 | 0.453 |
| PM | 0.116 | 0.070 | 0.842 | 0.137 | 0.084 | 0.775 | 0.174 | 0.110 | 0.650 | 0.188 | 0.122 | 0.600 | 0.221 | 0.145 | 0.412 |
| XGBoost | 0.115 | 0.076 | 0.845 | 0.129 | 0.084 | 0.799 | 0.161 | 0.109 | 0.699 | 0.177 | 0.127 | 0.645 | 0.200 | 0.142 | 0.519 |
| SVR | 0.144 | 0.100 | 0.757 | 0.165 | 0.116 | 0.673 | 0.215 | 0.150 | 0.462 | 0.240 | 0.167 | 0.345 | 0.252 | 0.178 | 0.238 |
| LSTM | 0.111 | 0.072 | 0.856 | 0.125 | 0.082 | 0.811 | 0.157 | 0.106 | 0.712 | 0.174 | 0.125 | 0.655 | 0.183 | 0.132 | 0.596 |
| MLP | 0.113 | 0.075 | 0.850 | 0.129 | 0.083 | 0.800 | 0.164 | 0.109 | 0.688 | 0.177 | 0.125 | 0.646 | 0.198 | 0.140 | 0.530 |

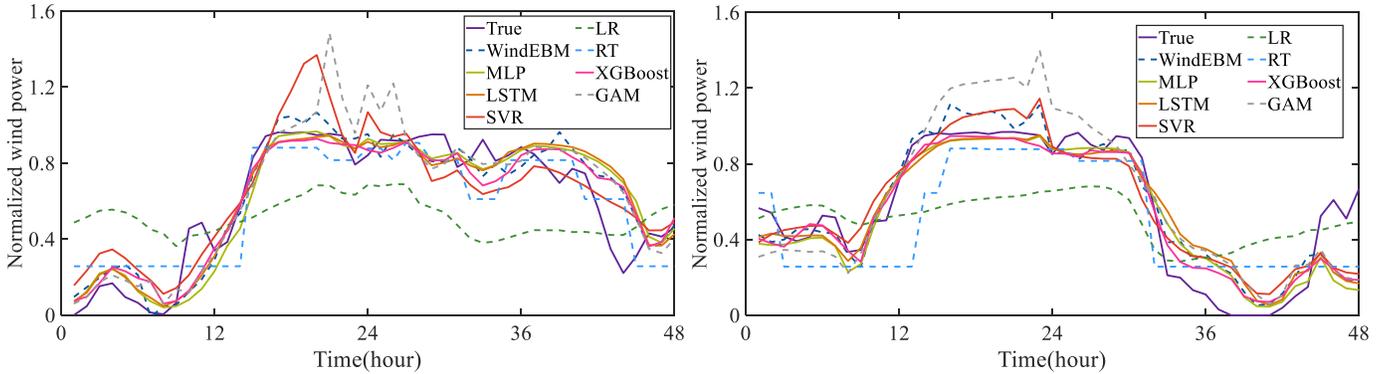

(a) Forecasted and real values from Sept. 22, 2013 to Sept. 23, 2013   (b) Forecasted and real values from Oct. 4, 2013 to Oct. 5, 2013

Fig. 5. The visualization of partial forecasted results for dataset 1, GEFCom 2014.

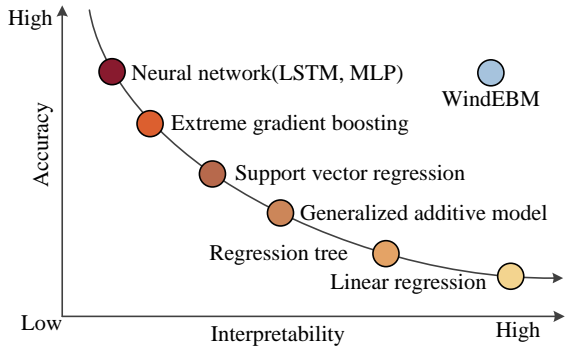

Fig. 6. The accuracy and interpretability of WindEBM and benchmarks in wind power forecasting.

Fig. 6 provides a visual insight into the accuracy and interpretability of the proposed WindEBM in comparison with benchmark models, using dataset 5 from the GEFCOM2014. Generally, this dual strength of transparency and high accuracy positions the proposed WindEBM as a compelling choice for reliable wind power forecasting.

### E. Time-Efficiency

To analyze the time efficiency of the proposed WindEBM and benchmark models, dataset 1 derived from the GEFCOM 2014 is used as a simple example. Then, the average training time and inference time of each model are presented in Table V.

Note the PM considers the most recent historical points as the forecasted values, resulting in both training and inference times of zero. Consequently, the PM is excluded from Table V.

TABLE V
TRAINING TIME AND INFERENCE TIME

| Time | WindEBM | LR | RT | GAM | XGBoost | SVR | LSTM | MLP |
|---|---|---|---|---|---|---|---|---|
| Training time(s) | 2.59 | 0.18 | 0.16 | 0.44 | 51.51 | 29.57 | 78.49 | 14.27 |
| Inference time(s) | 0.001 | 0.011 | 0.002 | 0.018 | 0.005 | 0.106 | 0.433 | 0.131 |

From Table V, the glass-box models (e.g., WindEBM, LR, RT, and GAM) demonstrate notably shorter training times (ranging from 0.16 to 2.59 seconds) compared to black-box models (e.g., XGBoost, SVR, LSTM, MLP) with training times spanning 14.27 to 78.49 seconds. Glass-box models also exhibit faster inference times (ranging from 0.001 to 0.018 seconds), showcasing their efficiency advantage over black-box counterparts (0.005 to 0.433 seconds).

This highlights the advantage of the proposed WindEBM in terms of time efficiency, making it suitable for real-time applications where fast wind power forecasting is crucial.

*F. Interpretability of Wind Power Forecasting*

In this section, dataset 1 from the GEFCOM2014 is used to interpret wind power forecasting from a global perspective and an instance perspective, respectively. Other datasets can be interpreted similarly.

**1) Global Interpretation**

Global interpretation helps to construct feature engineering, i.e., to understand which input features the model considers important for wind power forecasting. To display a global explanation, the training set is used to calculate the average absolute scores (i.e., the output of shape function in Eq. (5).) of input features and their interaction terms. Fig. 7 shows the average importance of each term in wind power forecasting.

From Fig. 7, U100 and V100 dominate the effect on wind power forecasting, followed by U10 and V10, and their interaction terms. With global interpretation, users can easily understand how important individual features are for wind power forecasting, and then can remove unnecessary features.

To evaluate the reliability of the global explanations provided by WindEBM, we also employed widely-used techniques, namely partial dependence plot (PDP) in [25] and permutation feature importance (PFI) in [26], to obtain the average importance of each feature, as shown in Fig. 8 and Fig. 9.

By comparing the global explanations provided by WindEBM with those from PDP and PFI techniques, we found them to be consistent (i.e., the ranking of feature importance is the same), which demonstrates the reliability of the global explanations provided by WindEBM. Furthermore, the global explanation provided by WindEBM is more comprehensive than that provided by PDP and PFI because it includes not only the importance of individual features, but also the importance of interaction terms.

In addition to observing the average importance of input features and their interaction terms, users can also quantitatively estimate the contribution of each term to wind power forecasting, because Eq. (5) is an additive model. This can be achieved by visualizing shape functions, as shown in Fig. 10 and Fig. 11.

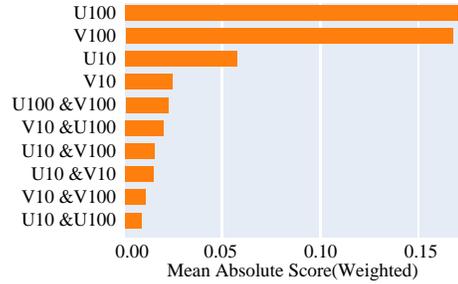

U:represents the wind speed in an east-west direction   10: wind speed at 10 meters
V:represents the wind speed in a north-south direction   100: wind speed at 100 meters
Fig. 7. Global interpretation provided by WindEBM: The average importance of input features in wind power forecasting.

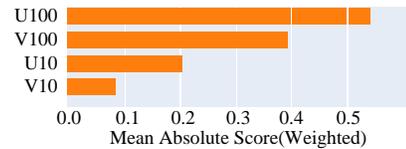

U:represents the wind speed in an east-west direction   10: wind speed at 10 meters
V:represents the wind speed in a north-south direction   100: wind speed at 10 meters
Fig. 8. Global interpretation provided by the PFI technique in [25]: The average importance of input features in wind power forecasting.

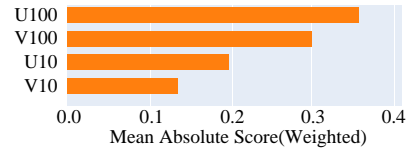

U:represents the wind speed in an east-west direction   10: wind speed at 10 meters
V:represents the wind speed in a north-south direction   100: wind speed at 10 meters
Fig. 9. Global interpretation provided by the PDP technique in [26]: The average importance of input features in wind power forecasting.

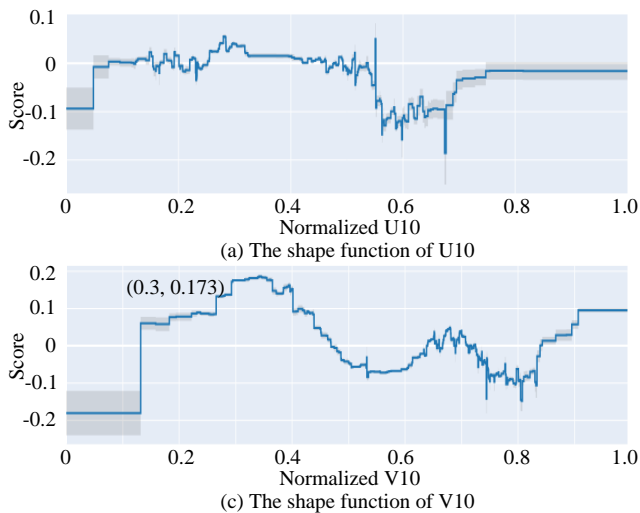
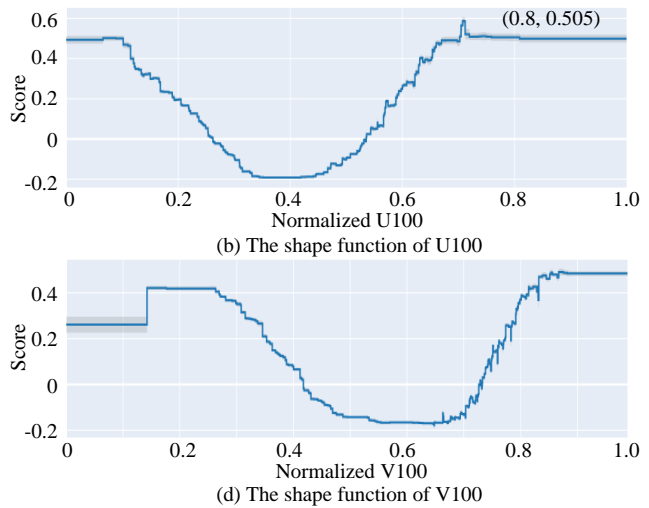

Fig. 10. The shape functions of wind speeds and wind directions.

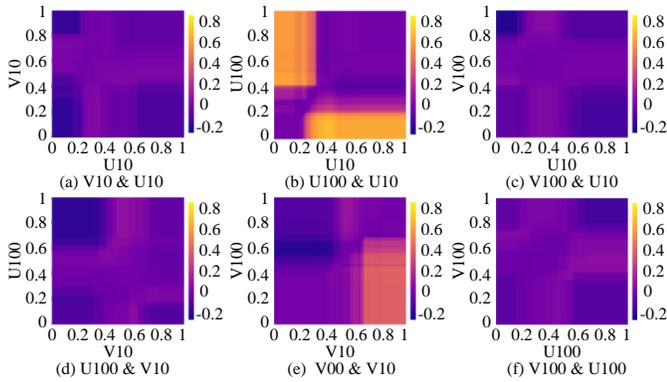

Fig. 11. The shape functions of interaction terms between U and V.

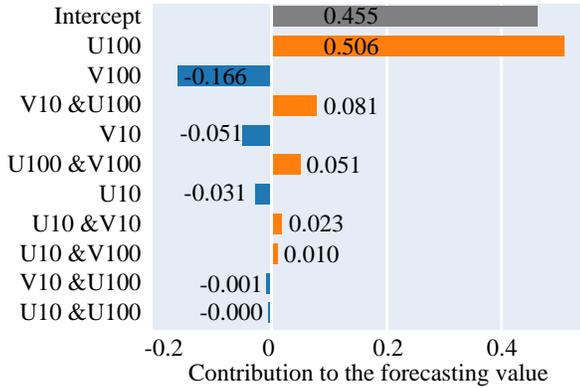

Actual(0.994),Forecasts(0.877)=0.455+0.506-······-0.001-0.000;

Fig. 12. The contribution of each input feature to the forecasted value.

The interpretation of these 1-D shape functions is that for a new observation with the normalized U100 of 0.8, WindEBM adds approximately +0.505 units to the forecasted value, as shown in Fig.10 (b). Similarly, for an alternative observation with the normalized V10 of 0.3, WindEBM would add about 0.173 units to the forecast, as presented in Fig.10 (c).

For the interpretability of 2-D shape functions, we can also measure the contribution of interaction terms to forecasted values. For instance, Fig. 11(a) presents a heatmap of a 2-D shape function, with U10 on the x-axis and V10 on the y-axis. When U10 is 0.4 and V10 is 0.6, the output of this 2-D shape function is approximately 0.15. This indicates the contribution of the interaction term formed by U10 and V10 to the forecasted value.

**2) Instance Interpretation**

Beyond the global interpretation, WindEBM also provides an instance interpretation to view the full breakdown of wind power forecasting given a specific sample. As an example, a sample is randomly selected from the test set, and then the inputs are fed into shape functions to get their contributions to the forecasted value, as shown in Fig. 12.

For a given sample, the shape functions can be considered are lookup tables. The contributions of input features and the intercept are summed to obtain the forecasted value. For this randomly selected sample, U100 and V100 dominate the effect on wind power forecasting, followed by the interaction term between U10 and V10. From Fig. 12, it is easy to understand why the forecasted value of the WindEBM for this sample is 0.877. This helps to identify and correct errors, improving the accuracy and reliability of wind power forecasting.

## V. CONCLUSION

Machine learning models achieve high prediction accuracy in wind power forecasting, but they are usually regarded as black-box models that lack interpretability. To address this issue, this paper proposes a glass-box approach with high accuracy for wind power forecasting. The following conclusions can be drawn from the simulation and analysis on seven real wind power datasets:

1) The proposed glass-box approach outperforms most benchmark models (e.g., LR, RT, PM, GAM, XGBoost, SVR, and MLP) and exhibits comparable performance to the top-performing neural networks (e.g., LSTM). This dual strength of transparency and high accuracy positions the proposed glass-box approach as a compelling choice for reliable wind power forecasting.

2) Compared with black-box models (e.g., XGBoost, SVR, LSTM, MLP), the proposed glass-box approach demonstrates notably shorter training time and inference time, making it suitable for real-time applications where fast wind power forecasting is crucial.

3) Global interpretation of the proposed glass-box approach helps to construct feature engineering, i.e., to understand which input features the model considers important for wind power forecasting. Besides, the instance interpretation of the proposed glass-box approach provides a way to view the full breakdown of wind power forecasting given a specific sample, which helps to identify and correct errors, improving the accuracy and reliability of wind power forecasting.

For future work, the effectiveness of the proposed glass-box approach on larger wind power datasets could be investigated as a primary consideration. Furthermore, this paper only investigates wind power forecasting within a 24-hour time horizon. Subsequent research could explore the applicability of the proposed method to wind power forecasting with a longer time horizon.

## ACKNOWLEDGMENT

The authors would like to thank Sophie Bosse and BKW FMB Energie AG for providing the Juvent dataset.